# Cross Knowledge-based Generative Zero-Shot Learning Approach with Taxonomy Regularization

Cheng Xie, Hongxin Xiang, Ting Zeng, Yun Yang*, Beibei Yu and Qing Liu

ABSTRACT

Although zero-shot learning (ZSL) has an inferential capability of recognizing new classes that have never been seen before, it always faces two fundamental challenges of the cross modality and cross-domain challenges. In order to alleviate these problems, we develop a generative network-based ZSL approach equipped with the proposed Cross Knowledge Learning (CKL) scheme and Taxonomy Regularization (TR). In our approach, the semantic features are taken as inputs, and the output is the synthesized visual features generated from the corresponding semantic features. CKL enables more relevant semantic features to be trained for semantic-to-visual feature embedding in ZSL, while Taxonomy Regularization (TR) significantly improves the intersections with unseen images with more generalized visual features generated from generative network. Extensive experiments on several benchmark datasets (i.e., AwA1, AwA2, CUB, NAB and aPY) show that our approach is superior to these state-of-the-art methods in terms of ZSL image classification and retrieval.

## 1. Introduction

Recently, as the rapid explosion of data scales, deep learning models have achieved state-of-the-art performance across a wide range of computer vision tasks such as image classification task [48, 52]. Unfortunately, in many applications, it is expensive and even impossible to collect enough labeled object categories [61, 63, 27]. To overcome this limitation, zero-shot learning (ZSL) [43, 29, 38, 59, 5, 8, 32] is proposed to establish such an learning system that can identify novel categories in the test stage, while there are no such categories during the training stage. In fact, ZSL is motivated by the human beings' cognitive learning mechanism in identifying unknown classes and has received increasing attention in recent years.

Generative ZSL attempts to learn the relationships between semantic features and visual features from seen categories and then generates synthesized images for unseen categories. However, semantic-to-visual feature learning faces two challenges, cross-modality, and cross-domain learning problems, that affect the final performance. Recently, state-of-the-art works are trying to reduce the adverse effects of cross-modality and cross-domain learning problems. Nevertheless, it is still challenging:

*The cross-modality challenge*: The relationships between visual features and the semantic features are exploited by a visual-semantic embedding either from the visual space to semantic space [66, 24, 60], or vice versa [67, 25, 40], or a shared common space [4, 37, 47]. However, visual-semantic embedding is faced with the problem of cross-modality, which leads to incomplete expression of semantic features and visual features in embedding, especially in two very similar categories where there is no difference in embedding space, as an example shown in Fig.1 (a). Therefore, Many effec-tive methods have been proposed to improve the expression of semantic and visual features, including attention-based methods [62, 68], knowledge graph-based methods [30, 55, 65, 36] and textual fusion-based methods [13]. Specifically, the current state-of-the-art textual fusion-based method [13] employs an elaborated strategy to merge textual descriptions of multiple categories. Although the method enriches the text representation of each category, it needs a lot of attempts or tuning parameters to fuse the "good" text. At the same time, there is a lot of noise in Wikipedia text, which leads to the increase of the text noise after fusion.

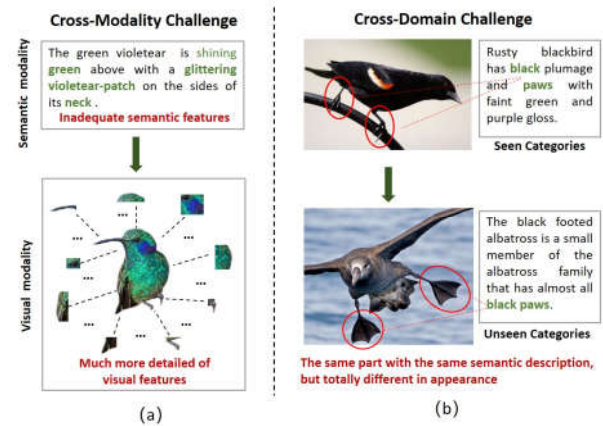

**Figure 1:** (a) An example about the cross-modality challenge; (b) An example about the cross-domain challenge.

*The cross-domain challenge*: There is always a cross-domain problem in ZSL since the seen categories might have few (or no) intersections with unseen categories. The visual appearance of the same attributes or textual descriptions can be obviously different in unseen class. E.g., in Fig.1 (b), the appearance of a particular part of a bird trained in seen categories might be significantly different from the appearance of the same part in unseen categories, even they share the

All the authors are with School of Software, Yunnan University, Kunming, 650504, China.

Cheng Xie and Hongxin Xiang contributed equally to this work.

*Corresponding author

ORCID(s): 0000-0002-9893-3436 (Y. Yang)

very similar semantic features ("both The black-footed albatross and Rusty blackbird have black paws."). It is also called the projection domain shift problem [31, 19] in ZSL. Many transductive ZSL approaches [63, 19, 20, 53] are proposed to alleviate this bias, that is, the unseen categories data and the seen categories data are aggregated together to learn more generalized visual-semantic embedding to adapt unseen categories. However, it is hard to obtain the visual features of the unseen class from the first place. The applicability of these methods is thus inadequate in most ZSL cases.

In this paper, to address the above challenges, we propose a novel generative Zero-Shot learning approach. The proposed approach meets the cross-modality challenge by applying Cross-Knowledge Learning (CKL) on seen categories. CKL uses taxonomic knowledge to enhance the semantic-to-visual feature embedding on family-level, genus-level and species-level, respectively. For each seen image, all the corresponding knowledge extracted from taxonomy is applied to the semantic-to-visual feature embedding process. It significantly enriches the semantic features for each seen image that alleviate inadequate semantics problem in cross-modality learning. Moreover, the proposed approach meets the cross-domain challenge by applying the term of Taxonomy Regularization (TR) to the loss function. TR leads to more generalized visual features to be generated for synthesized images that tend to intersect with unseen images. The adverse effects from cross-domain decrease while the visual feature intersections between synthesized images and unseen images increase. To sum up, the main contributions of the paper can be summarized as follows:

- We propose the Cross Knowledge Learning (CKL) that enables more relevant semantic features to be trained for semantic-to-visual feature embedding in ZSL. It obviously enriches the semantic features during the cross-modality learning.

- We propose the Taxonomy Regularization (TR) that leads more generalized visual features to be generated to increase the intersections with unseen images in ZSL. It significantly alleviates the adverse effects from cross-domain learning.

- Our approach consistently outperforms state-of-the-art methods of zero-shot recognition, zero-shot retrieval, and generalized zero-shot learning on six major benchmarks for ZSL task.

## 2. Related Works

### 2.1. Early Zero-Shot Learning(ZSL) Approaches

A key idea to facilitate zero-shot learning is to find a common semantic representation that seen and unseen classes can share. Attributes and textual descriptions are shown to be effective shared semantic representations, allowing knowledge to be transferred from seen to unseen classes. As one of the pioneering works, a Direct Attribute Prediction (DAP) approach [34] is proposed, which assumes that the attributes are independent and estimates the posterior of the unseen class based on the attribute prediction probabilities. Without the independent assumption, the Attribute Label Embedding (ALE) approach [1] is proposed to consider attributes as the semantic embedding of classes and thus tackle ZSL as a linear joint visual-semantic embedding. Most of these early ZSL methods are based on semantic embedding, and they are easy to ignore a large number of discriminative visual information when semantic embedding. Our method uses semantic-to-visual embedding, which can retain more discriminative visual features.

### 2.2. Semantic-to-Visual Learning for ZSL

Recently, many ZSL methods [59, 66, 24, 60, 67, 25, 40, 13, 28, 26] are proposed to facilitate zero-shot learning to establish a mapping from semantic space to visual space. The motivation is to alleviate the hubness problem that commonly suffered by nearest neighbour search in a high dimensional space [45]. There is a similar idea in recent methods, which constructed a deep model that took visual features extracted by CNN [21, 54] and semantic features extracted by word embedding [49] as input, and trained the model with objective that the visual and semantic features of the same class would be well aligned under linear transformations. Based on this idea, the approach [28] designs an effective strategy that exploits the attribute to guide the generation of visual hash codes for cross-modal retrieval in task; the approach [26] presents a deep regression model to project the visual features into the semantic space for zero-shot multi-label classification. Our approach uses taxonomy information to learn more generalized model, which makes the synthesized visual features to tend to intersect with unseen images.

### 2.3. Generative ZSL Approaches

A typical representative of semantic-to-visual learning is the generative ZSL approaches [59, 67, 15], which use Generative Adversarial Network (GAN) as the basic structure of deep model and reduce ZSL to a conventional classification task by generating fake samples. Earlier approaches assumed that the visual space of each class had a Gaussian distribution, while the probability densities of the unseen classes were modeled as a linear combination of the seen classes distributions [6]. Instead, the paper [39] proposed a one-to-one mapping approach that limits the synthesized examples. Recently, approaches [59, 67] were the pioneer to combine Generative Adversarial Networks (GANs) [42, 10] to generate fake visual features from the textual descriptions of unseen classes. Different from ACGAN [42], the approach [67] added a visual pivot regularizer (VPG) that encourages generations of each class to be close to the average of its corresponding real visual features. In order to alleviate the problem of insufficient semantics, approaches [23, 17, 33] transform the synthesized visual samples back into the semantic space, accompanies with cycle consistency loss; approach [13] proposes a creative mechanism to randomly synthesize more texts to enhance semantic representation; approaches [23, 41, 25] use semantic-to-visual, visual-to-semantic multiple mapping to retain enough semantic in-formation to further ensure semantic-visual interaction. Although these GAN-based methods are a powerful generative models, they did not explicitly guide the model to generate novel content beyond the seen data. Our approach is also a generative model. The difference from the existing models is through the addition of TR loss that encourage the model to explore the more generalized space of visual generation.

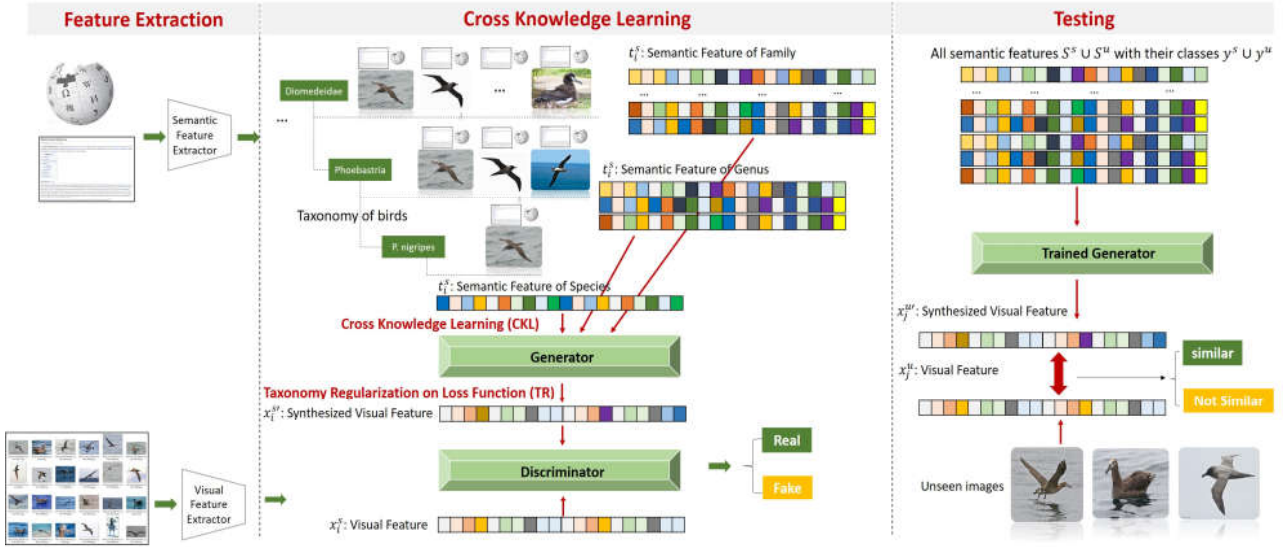

**Figure 2:** The overall framework of the proposed approach.

### 2.4. Semantic Representations

Additional information enables the implementation of ZSL, for example, the attributes or textual description of the unseen classes. It has achieved considerable progress in studying attribute representation [34, 1, 35, 64]. Attributes are a collection of semantic characteristics that uniquely describe unseen classes. Attribute-based methods are first proposed in [35], which train a classifier for each attribute. Another trend in ZSL is to use online textual descriptions [15, 12, 44, 46]. Textual descriptions can be easily extracted from online resources such as Wikipedia with minimal overhead, eliminating the need to define hundreds of attributes and populate them for each class/image. Our approach achieves the zero-shot recognition task based on attributes and online textual description, Wikipedia articles. Unlike existing, our approach not only provides richer semantic features to train semantic-to-visual embedding by using taxonomic knowledge, but also makes the model generate more generalized visual features, which tend to intersect with unseen images.

## 3. The Proposed Approach

### 3.1. Problem Definition

Let $\mathcal{Y}$ denotes the set of all classes. $y^s \in \mathcal{Y}$ and $y^u \in \mathcal{Y}$ represent the seen (training) class and unseen (testing) class respectively, and $y^s \bigcap y^u = \phi$. The visual features of seen and unseen classes are represented as $x_i^s = \psi(X_c^s) \in \mathcal{V}^s$ and $x_i^u = \psi(X_c^u) \in \mathcal{V}^u$, where $X_c^u$ and $X_c^s$ represent images (samples) of the $c$-th seen class and unseen class respectively, and $\mathcal{V}^s \bigcap \mathcal{V}^u \neq \phi$. $\psi(*)$ is a feature extraction function to transform an image into visual features $\mathcal{V}$. The semantic features of seen and unseen classes are represented as $t_i^s = \varphi(T_c^s) \in S^s$ and $t_i^u = \varphi(T_c^u) \in S^u$, where $T_c^u$ and $T_c^s$ represent the text or attribute of the $c$-th seen class and unseen class respectively, and $S^s \bigcap S^u \neq \phi$. $\varphi(*)$ is a word embedding function to map text or attribute into semantic space $S$. A training data points can be defined as:

$$\mathcal{D}^s = \{(x_i^s, y_i^s, t_i^s), i = 1, ..., N^s\} \qquad (1)$$

where $N^s$ is the number of samples from seen classes. A testing data point can be also defined as:

$$\mathcal{D}^u = \{(x_j^u, y_j^u, S^s \cup S^u \to \mathcal{V}^s \cup \mathcal{V}'^u), j = 1, ..., N^u\} \quad (2)$$

where $N^u$ is the number of samples from unseen classes. $y_j^u$ is the corresponding class of $x_j^u$. $\mathcal{V}'^u$ is the synthesized unseen visual features generated from the corresponding unseen semantic features. Thus, the task of zero-shot image classification is to give an unseen visual feature $x_j^u$, the model transforms all the semantic features into visual features, including seen and unseen features($S^s \cup S^u$). Then, the model compares the unseen visual feature $x_j^u$ with all synthesized visual features (generated from $S^s \cup S^u$). Finally, the model selects the most matched synthesized visual features, and get the corresponding class $y_j^u$ as the result.

### 3.2. The Model Architecture

As showed in Fig. 2, the skeleton of the model is a Generative Zero-Short Learning Network (G-ZSL). In this work, three state-of-art G-ZSL networks (ACGAN [42], GAZSL [67], and FeatGen [59]) are selected as the skeleton of the model. These three G-ZSL networks are also the baselines, named Baseline17, Baseline18, and Baseline18-2, respectively, compared with the proposed model. The generator $G$ and discriminator $D$ are represented as $\mathbb{R}^S \times \mathbb{R}^M \to \mathbb{R}^V$, $\mathbb{R}^V \to \{0,1\} \times \mathbb{L}_{cls}$, where $\mathbb{R}^M$ represents the mapping from semantic features to visual features, and $\mathbb{L}_{cls}$ represents the labels $y^s \in \mathcal{Y}$ corresponding to the visual features of the seen classes. The model uses the following steps to train the generative zero-shot learning network:

- *Feature Extraction*: The model extracts fine-grained visual features by using the part-based FC layer of VPDE-net [64]. The model extracts semantic features by using Term Frequency Inverse Document Frequency (TF-IDF) [49] according to the related works [15, 12].

- *Cross Knowledge Learning (CKL)*: For a given training data point $(x_i^s, y_i^s, t_i^s)$, CKL enriches the semantic feature $(t_i^s)$ for each visual feature $(x_i^s)$ by extracting extra text or attributes from cross classes through the taxonomy.

- *Taxonomy Regularization (TR)*: TR first calculates the center point of the visual feature of family, genus, and species according to the Taxonomy. TR is then applied on $G'$s loss function to force the synthesized visual features to close to the center point of family, genus, and species.

- *Training and Testing:* The main purpose of training a generative ZSL model is to learn the semantic-to-visual feature mappings by $G$. Then, during the testing, the trained $G$ can generate synthesized visual features from semantic features to match unseen visual features.

According to the model architecture, section 3.3 introduces CKL in detail. Section 3.4 explain the implementation of TR. Section 3.5 describes the training and testing process of the model.

### 3.3. Cross Knowledge Learning (CKL)

Previous zero-shot learning methods normally use the single semantic knowledge to learn the semantic-to-visual feature embedding. It inevitably increases cross-modality problems, since, in the practice, visual features are much more richer and detailed than the corresponding semantic features. It also significantly increases the imbalance between semantic features and visual features. In this work, beyond the single semantic knowledge, Cross Knowledge Learning (CKL) is trying to enrich the semantic features by using multiple semantic knowledge. The idea of CKL is that given a seen class $y_i^s$, we first use its corresponding semantic features $t_i^s$ to train the semantic-to-visual feature embedding. Then, more semantic features $\{ \hat{t}_{i,1}^s, \hat{t}_{i,2}^s, ..., \hat{t}_{i,n}^s \}$ of $y_i^s$'s similar classes $\{ \hat{y}_{i,1}^s, \hat{y}_{i,2}^s, ..., \hat{y}_{i,n}^s \}$ (classes are considered to be similar if they share the same category in the taxonomy) are also applied to train the semantic-to-visual feature embedding. Thus, in Cross Knowledge Learning, the training data

point can be extended as following:

$$\mathcal{D}^s = (x_i^s, y_i^s, \{t_i^s, \bigcup_{k=1}^{K} \hat{t}_{i,k}^s\}), i = 1, ..., N^s \quad (3)$$

where $K$ is the total number of $y_i^s$'s similar classes in taxonomy. Compared with the training data point of the traditional ZSL learning (Eq. 1), our learning model uses more relevant knowledge ($t_i^s \cup \bigcup_{k=1}^{K} \hat{t}_{i,k}^s$ versus single $t_i^s$) to learn the semantic-to-visual feature embedding. It can be expected to reduce the imbalance between semantic features and visual features. Furthermore, in the work, three taxonomy categories (Species, Genus and Family) are used. Eq. 3 can be further divided into three data points according to Species, Genus and Family category of the taxonomy (See 4.1 for details):

$$\mathcal{D}^{s,Species} = (x_i^s, y_i^s, t_i^s), i = 1, ..., N^s \quad (4)$$

$$\mathcal{D}^{s,Genus} = (x_i^s, y_i^s, \{t_i^s, \bigcup_{k=1}^{Gen} \hat{t}_{i,k}^s\}), i = 1, ..., N^s \quad (5)$$

$$\mathcal{D}^{s,Famlily} = (x_i^s, y_i^s, \{t_i^s, \bigcup_{k=1}^{Fam} \hat{t}_{i,k}^s\}), i = 1, ..., N^s \quad (6)$$

where $Gen$ and $Fam$ represent the total number of $y_i^s$'s similar classes in Genus and Family categories of the taxonomy respectively.

### 3.4. Taxonomy Regularization (TR)

In the Generative Zero-Shot Learning (G-ZSL), the generator is used to generate synthesized visual features to match (predict) the unseen visual features. To further alleviate the cross-domain problem, Taxonomy Regularization (TR) is proposed to regulate the generator to produce more generalized samples. The intuition of TR is that the unseen features can not be fully learned from the seen features since the seen classes and unseen classes are not intersected (i.e., $y^s \bigcap y^u = \phi$). It implies the more features learned only from a single class, the fewer interactions between synthesized features and unseen features can be obtained. Thus, the idea of TR is to force the generator to produce synthesized visual features that deviate from the corresponding seen classes. Meanwhile, TR also guides the generator to produce more synthesized visual features that close to the similar classes (classes are considered to be similar if they share the same category in the taxonomy) according to the categories of the taxonomy. It reasonably increases the diversity of the generator to produce various synthesized visual features that can be expected to intersect with the unseen features. In this way, TR loss function is thus designed to accomplish the idea:

$$L_{tr} = \lambda_s L_{tr}^{species} + \lambda_g L_{tr}^{genus} + \lambda_f L_{tr}^{family} \quad (7)$$

where, $\lambda_s$, $\lambda_g$ and $\lambda_f$ are the weights of the regularized items, and are constrained to $\lambda_s + \lambda_g + \lambda_f = 1$. $L_{tr}^{species}$, $L_{tr}^{genus}$ and $L_{tr}^{family}$ are defined as:

$$L_{tr}^{species} = \frac{1}{N^s} \sum_{i=1}^{N^s} \parallel G(t_i^s, z) - \frac{1}{N_i^{spec}} \sum_{k=1}^{N_i^{spec}} x_k^s \parallel^2$$

$$L_{tr}^{genus} = \frac{1}{N^s} \sum_{i=1}^{N^s} \parallel G(t_i^s, z) - \frac{1}{N_i^{gen}} \sum_{k=1}^{N_i^{gen}} x_k^s \parallel^2 \quad (8)$$

$$L_{tr}^{family} = \frac{1}{N^s} \sum_{i=1}^{N^s} \parallel G(t_i^s, z) - \frac{1}{N_i^{fam}} \sum_{k=1}^{N_i^{fam}} x_k^s \parallel^2$$

where $G(t_i^s, z)$ is the synthesized visual features of class $y_i^s$ produced by $G$ with a random noise $z$. $(\frac{1}{N_i^{spec}} \sum_{k=1}^{N_i^{spec}} x_k^s)$ is the visual feature center of all classes belong to $y_i^s$'s Species. Similarly, we can have the visual feature centers of $y_i^s$'s Genus and Family. Eq. 7 and 8 force synthesized visual features close to the centers of Species, Genus and Family, not to some particular samples. It enhances the generalization ability of the generator that helps the model to produce more various visual features. After, we combine TR loss $L_{tr}$ into the final loss function of the generator as follows:

$$L_G = -\mathbb{E}[D(G(t^s, z))] + L_{cls}(G(t^s, z)) + L_{tr}(G(t^s, z) \quad (9)$$

where the first term is Wasserstein loss [3] and the second term is additional classification loss [67] corresponding to class labels. $D$ is the Discriminator of the model. The loss function of $D$ is the same with the previous work [67].

### 3.5. Learning Scheme

Taking the data points of CKL ($D^{s,Species}$, $D^{s,Genus}$ and $D^{s,Family}$) as the input, TR as the new term on the loss function, the learning scheme of the proposed model is presented in Algorithm 1.

Given any semantic features $t^u$ $from$ $S^u$, synthesized visual features can be generated $x_j^{u\prime}$ from $G(t_j^u, z)$ after the model was learned. For a semantic feature $t_j^u$, various visual features can be synthesized by repeatedly sampling $z$. Taking all the semantic features of unseen class as input, the model can generate all synthesized visual features for unseen class. Given an unseen image, the model calculates similarity between the real visual feature $x_j^u$ and synthesized visual features $x_j^{u\prime}$. The class label of an unseen image can be determined by finding the label $y_j^u$ corresponding to the synthesized visual feature $x_j^{u\prime}$ that is the most similar to the real visual feature $x_j^u$.

## 4. Experiments

Three generative ZSL methods, Baseline17 (ACGAN) [42], Baseline18 (GAZSL) [67] and Baseline18-2 (FeatGen) [59], are selected as the baseline methods in the experiment.

---

**Algorithm 1: Cross-Knowledge Learning**

**Input:** $D^{s,Species}$, $D^{s,Genus}$ and $D^{s,Family}$

1 **for** *each Species, Genus, and Famlily* **do**
2    Caculating the visual feature center $\bar{x}^{spec}$, $\bar{x}^{gen}$, and $\bar{x}^{fam}$.
3 **end**
4 **while** *iteration−−* **do**
5    **for** *each $x_i^s$ and $t_i^s$ in $D^{s,Species}$* **do**
6      $x_i^{s\prime} \leftarrow G(t_i^s, z)$
7      Caculating the loss through $x_i^s$ and $x_i^{s\prime}$.
8      Updating the Discriminator $D$.
9      $L_{tr}^{Species} += \|G(t_i^s \cup \{i_i^s\}, z) - \bar{x}_i^{spec}\|^2$
10    **end**
11    **for** *each $x_i^s$ and $t_i^s \cup \{i_i^s\}$ in $D^{s,Genus}$* **do**
12      $L_{tr}^{genus} += \|G(t_i^s \cup \{i_i^s\}, z) - \bar{x}_i^{gen}\|^2$
13    **end**
14    **for** *each $x_i^s$ and $t_i^s \cup \{i_i^s\}$ in $D^{s,Famlily}$* **do**
15      $L_{tr}^{Family} += \|G(t_i^s \cup \{i_i^s\}, z) - \bar{x}_i^{fam}\|^2$
16    **end**
17    $L_{tr} = \frac{\lambda_s}{|D|} L_{tr}^{species} + \frac{\lambda_g}{|D|} L_{tr}^{genus} + \frac{\lambda_f}{|D|} L_{tr}^{family}$
18    Caculating the final loss of $G$ from Eq. 9.
19    Updating the Generator $G$.
20 **end**

---

**Table 1**
Statistical information about the datasets.

| Dataset | $|S|$ | $S_{type}$ | $|\mathcal{X}|$ | $|\mathcal{Y}|$ | $|\mathcal{Y}^s|$ | $|\mathcal{Y}^u|$ |
|---|---|---|---|---|---|---|
| CUB(easy)[56] | 7551 | Wiki | 11788 | 200 | 120+30 | 50 |
| CUB(hard)[56] | 7551 | Wiki | 11788 | 200 | 130+30 | 40 |
| NAB(easy)[22] | 13217 | Wiki | 48562 | 404 | 200+123 | 81 |
| NAB(hard)[22] | 13217 | Wiki | 48562 | 404 | 200+123 | 81 |
| CUB[56] | 312 | Y | 11788 | 200 | 100+50 | 50 |
| AwA1[58] | 85 | Y | 30475 | 50 | 27+13 | 10 |
| AwA2[58] | 85 | Attr | 37322 | 50 | 30+10 | 10 |
| aPY[16] | 64 | Attr | 15339 | 32 | 15+5 | 12 |

The terms Baseline+CKL, Baseline+TR, and Baseline+CKL+TR represent our method with Taxonomy Regularization (TR) and Cross-Knowledge Learning (CKL) combinations.

### 4.1. Experimental Settings

*Datasets:* The proposed approaches are evaluated on the four most commonly used benchmarks in ZSL field: Caltech-UCSD-Birds 200-2011 (CUB) [56], North America Birds (NAB) [22], Animals with Attributes 2 (AwA2) [58] and attributes Pascal and Yahoo (aPY) [16]. Table.1 shows the statistical information about these datasets.

In Table.1, $|S|$ represents the dimension of semantic features. $S_{type}$ denotes the type of semantic features. $|\mathcal{X}|$ represents the number of images. $|\mathcal{Y}|$ represents the number of classes in seen(training)+unseen(test), $|\mathcal{Y}^s|$ represents the number of seen classes, and $|\mathcal{Y}^u|$ represents the number of unseen classes.

Especially, AwA2 and aPY have provided semantic features by adding attribute annotations for all samples. The datasets have also extracted the visual features for all samples by a pre-trained RestNet101 [21]. Thus, we can directly use the datasets without feature extraction. In CUB and NAB, we use Term Frequency-Inverse Document Frequency (TF-IDF) [49], and follow the extraction process of [15, 12], to extract semantic features from the corresponding Wikipedia pages. We extract fine-grained visual features of samples (images) by using the part-based FC layer of VPDE-net [64]. The images first are resized to $224 \times 224$ and fed forward to the VPDE-net. Then, a 512D feature vector for each detected image-part can be obtained from the VPDE-net. There are seven image-parts be detected in CUB dataset, which are head, back, belly, breast, leg, wing and tail. In NAB dataset, the "leg" part is missing because there is no annotation for this part in the original dataset. At last, the full dimensions of visual features in CUB and NAB are 3584D ($512D \times 7$ parts) and 3072D ($512D \times 6$ parts) respectively.

*Data Division:* We obtain the taxonomy information from https://encyclopedia.thefreedictionary.com/, as well as the class description from https://en.wikipedia.com. For each class in CUB, NAB, AwA2 and aPY datasets, the class label is searched in https://encyclopedia.thefreedictionary.com/ to get the corresponding Species. Then, the corresponding Family and Genus can also be obtained through the Family-Genus-Species hierarchy. At last, all samples in datasets are labeled by Family- Genus- Species for the three data points.

*Metrics:* In zero-shot learning, two popular metrics in ZSL are used to evaluate the performance of our approach: Top-1 unseen class accuracy for standard ZSL evaluation and the area under Seen-Unseen curve (AUSUC) for generalized ZSL evaluation. The Top-1 accuracy is widely used in many existing works [13, 15, 44, 58, 67], which is computed independently for each class before dividing their cumulative sum by the number of classes. However, the Top-1 accuracy is not enough to evaluate the generalization performance of the model on seen and unseen dataset. Therefore, the AUSUC metric, which is first proposed in [7], is used to measure the generalization ability of the model. The metric considers the model to classify images of both seen classes $y^s$ and unseen classes $y^u$ into $y = y^s \bigcup y^u$. The accuracy is denoted as $A_{y^s \rightarrow y}$ and $A_{y^u \rightarrow y}$ respectively. Then a balancing parameter $\lambda$ is introduced to draw Seen-Unseen accuracy curve (SUC) and Area Under SUC is used to measure the generalization performance of models in ZSL. In this paper, we map the semantic features of seen and unseen classes to the synthesized visual features $\mathcal{V}$, and $k$-Nearest-Neighbor (KNN) method is used to predict the class labels which the synthesized visual features belong to.

## 4.2. Experimental Results on CUB and NAB

There are two training-testing split strategies in CUB and NAB dataset, which are Super-Category-Shared (easy) and Super-Category-Exclusive Splitting (hard). The two split strategies are divided according to whether they share the same parent class. In the easy splitting, for each unseen class, there is one or more seen classes of the same parent class. For instance, the seen class "Indigo Bunting" and the unseen class "Lazuli Bunting" have the same parent class "Bunting". In the hard splitting, the unseen classes never share the same parent with seen classes. Obviously, seen class and unseen class have high correlations in the easy splitting, while seen class and unseen class have little correlation in the hard splitting. Therefore, zero-shot classification and retrieval are more difficult in Super-Category-Exclusive than in Super-Category-Shared.

### 4.2.1. Results of ZSL Image Classification

Table 2 shows the state-of-the-art comparisons on CUB and NAB with easy and hard splittings. The red and blue value represent the increasing and decreasing numbers compared with the corresponding baseline method respectively. The proposed method significantly outperforms Baseline17, Baseline18, Baseline18-2 and the state-of-the-art method (CIZSL [13]) in CUB and NAB on all tasks. Especially, in the hard splitting of CUB and NAB, the proposed method achieves 46.6% and 16.3% improvements on Top-1 Accuracy testing, 40.2% and 34.5% improvements on Seen-Unseen AUC testing, compared with Baseline18. Similar improvements are also achieved comparing with Baseline18-2. In the easy splitting of CUB and NAB, the proposed method obtains 4.8% and 3.4% improvements on Top-1 Accuracy testing, 13.5% and 9.3% improvements on Seen-Unseen AUC testing, compared with Baseline18. Similar improvements are also obtained comparing with Baseline18-2. Compared with the state-of-the-art methods, the proposed method achieves 1.0% improvement and a slight decrease (only 0.7% less) in the easy splitting, 4.9% and 11.8% improvements in the hard splitting, on Top-1 accuracy testing of CUB and NAB. Moreover, the proposed method achieves 2.2% and 2.3% improvements in the easy splitting, 3.2% decrease and 14.7% improvements in the hard splitting, on Seen-Unseen AUC testing of CUB and NAB.

### 4.2.2. Result of ZSL Image Retrieval

We use mean Average Precision (mAP) metric to evaluate the performance in the zero-shot retrieval task. The task is defined as retrieving images by using the semantic features (not visual features) of the unseen class. Table 3 shows the zero-shot retrieval results compared with state-of-the-art methods. The splitting strategies are retrieving 25%, 50% and 100% of the images of each class. In the detail, we first generate $n$ synthesized visual features $x_j^u$ from the corresponding semantic features $t_j^u$, and calculate the visual center $\tilde{x}^u = \frac{1}{n} \sum_{j=1}^{n} x_j^u$. Then the visual center $\tilde{x}^u$ is used to retrieve images according to the nearest neighbor strategy. Compared with Baseline17, the proposed method obtains in average 194.1% improvements in CUB, and in average 97.7% improvements in NAB. Compared with the other state-of-the-art method (ZSLPP[15]), the proposed method obtains in average 23.3% improvements in CUB, and in average 10.9% improvements in NAB. Compared with Baseline18, the proposed method obtains 7.6% average improve-ments in CUB, 5.1% average improvements in NAB.

**Table 2**
Result of ZSL Image Classification on **CUB** and **NAB**. We reproduce the results of GDAN, whose source code is available online.

| | Top-1 Accuracy | | | | Seen-Unseen AUC | | | |
|---|---|---|---|---|---|---|---|---|
| | CUB | | NAB | | CUB | | NAB | |
| methods | Easy | Hard | Easy | Hard | Easy | Hard | Easy | Hard |
| Baseline17 [42] | 15.7 | 6.6 | 12.1 | 4.1 | 3.1 | 3.0 | 2.1 | 0.2 |
| WAC-Linear[12] | 27.0 | 5.5 | - | - | 23.9 | 4.9 | 23.5 | - |
| WAC-Kernel[14] | 33.5 | 7.7 | 11.4 | 6.0 | 14.7 | 4.4 | 9.3 | 2.3 |
| ESZSL[47] | 28.5 | 7.4 | 24.3 | 6.3 | 18.5 | 4.5 | 9.2 | 2.9 |
| ZSLNS[44] | 29.1 | 7.3 | 24.5 | 6.8 | 14.7 | 4.4 | 9.3 | 2.3 |
| $SynC_{fast}$[6] | 28.0 | 8.6 | 18.4 | 3.8 | 13.1 | 4.0 | 2.7 | 3.5 |
| ZSLPP[15] | 37.2 | 9.7 | 30.3 | 8.1 | 30.4 | 6.1 | 12.6 | 3.5 |
| CIZSL[13] | 44.6 | 14.4 | 36.6 | 9.3 | 39.2 | 11.9 | 24.5 | 6.4 |
| CANZSL[9] | 45.8 | 14.3 | 38.1 | 8.9 | 40.2 | **12.5** | 25.6 | 6.8 |
| GDAN[23] | 44.2 | 13.7 | **38.3** | 8.7 | 38.7 | 10.9 | 24.1 | 5.9 |
| Baseline18 [67] | 43.7 | 10.3 | 35.6 | 8.6 | 35.4 | 8.7 | 20.4 | 5.8 |
| **Baseline18+CKL** | 43.9$^{+0.2}$ | 11.5$^{+1.2}$ | 35.7$^{+0.1}$ | 9.0$^{+0.4}$ | 38.6$^{+3.2}$ | 10.7$^{+2.0}$ | 22.7$^{+2.3}$ | 5.7$^{-0.1}$ |
| **Baseline18+CKL+TR** | 45.8$^{+2.1}$ | **15.1**$^{+4.8}$ | 36.8$^{+1.2}$ | 10.0$^{+1.4}$ | 40.2$^{+4.8}$ | 12.2$^{+3.5}$ | 22.3$^{+1.9}$ | 7.8$^{+2.0}$ |
| Baseline18-2 [59] | 43.9 | 9.8 | 36.2 | 8.7 | 34.1 | 7.4 | 21.3 | 5.6 |
| **Baseline18-2+CKL** | 44.0$^{+0.1}$ | 12.6$^{+2.8}$ | 36.8$^{+0.6}$ | 9.5$^{+0.8}$ | 39.4$^{+5.3}$ | 11.5$^{+4.1}$ | **26.2**$^{+4.9}$ | 6.2$^{+0.6}$ |
| **Baseline18-2+CKL+TR** | **46.3**$^{+2.4}$ | 14.4$^{+4.6}$ | 37.3$^{+1.1}$ | 10.4$^{+1.7}$ | **41.1**$^{+7.0}$ | 12.1$^{+4.7}$ | 24.9$^{+3.6}$ | **7.4**$^{+1.8}$ |

**Table 3**
Result of ZSL Image Retrieval.

| | CUB | | | NAB | | |
|---|---|---|---|---|---|---|
| methods | 25% | 50% | 100% | 25% | 50% | 100% |
| Baseline17 [42] | 18.0 | 17.5 | 15.2 | 21.7 | 20.3 | 16.6 |
| ESZSL[47] | 27.9 | 27.3 | 22.7 | 28.9 | 27.8 | 20.85 |
| ZSLNS[44] | 29.2 | 29.5 | 23.9 | 28.78 | 27.27 | 22.13 |
| ZSLPP[15] | 42.3 | 42.0 | 36.6 | 36.9 | 35.7 | 31.3 |
| CIZSL[13] | 50.3 | 48.9 | 46.2 | 41.0 | **40.2** | 34.2 |
| Baseline18 [67] | 49.7 | 48.3 | 40.3 | **41.6** | 37.8 | 31.0 |
| **Baseline18 +CKL** | 51.2$^{+1.5}$ | 49.5$^{+1.2}$ | 42.7$^{+2.4}$ | 40.8$^{-0.8}$ | 38.2$^{+0.4}$ | 33.2$^{+2.2}$ |
| **Baseline18 +CKL+TR** | **53.1**$^{+3.4}$ | **51.2**$^{+2.9}$ | **44.8**$^{+4.5}$ | 41.3$^{-0.3}$ | 38.7$^{+0.9}$ | **35.2**$^{+4.2}$ |

The representative ZSL retrieval results of the proposed method are shown in Fig. 3. In the figure, the first column is a sample image of the class to be retrieved (ground truth). The second to the sixth columns are the images be retrieved by the input semantic features. The order of these five columns represents the Top-5 matched images. The green box indicates the retrieved image is correct. The red box means the retrieved image is wrong. The experimental result shows the proposed model can effectively retrieve images from unseen classes. Specially, as shown in the third and last rows in Fig. 3, the mismatched images are very close to the ground truth (first column), even the human hard to distinguish these images. The synthesized visual features from semantic features are very close to the real visual features.

## 4.3. Experimental Results on CUB, AwA1, AwA2 and aPY

In order to evaluate that our approach is still effective under different semantic representations, GBU setting [57] is applied to change the semantic representation from texts of Wikipedia to attributes. The experiment is conducted on CUB[56], AwA1 [58], AwA2 [58] and aPY [16] dataset. As shown in Table 1, CUB consists of 11788 images of 200 birds with 312 attributes, AwA1 consists of 30475 images of 50 animal classes with 85 attributes, AwA2 consists of 37322 images of 50 animal classes with 85 attributes, and aPY consists of 15339 images of 32 objects classes with 64 attributes.

The result shows the proposed approach significantly outperforms other comparisons in CUB, AwA1, AwA2 and aPY dataset, as shown in Table 4. Compared with Baseline18*[1], a The result is selected from $\mathcal{L}_{cWGAN}^{S} + \mathcal{L}_{GM}$ in Table 3 of the GMN paper [50]. The result cannot be compared since it includes unseen samples in the training process.

---

[1]baseline18* is our implementation of baseline18 [67] that has a higher performance on harmonical mean. The main changes are: (1) baseline18* used visual features of seen classes extracted from real samples in testing;

**Table 4**
Result of ZSL Image Classification on **CUB**, **AwA1**, **AwA2** and **aPY** datasets. T1=Top-1 Accuracy, U=Unseen classes, S=Seen classes, H=harmonical mean (2*U*S/(U+S)).

| Methods | CUB | | | | AwA1 | | | | AwA2 | | | | aPY | | | |
|---|---|---|---|---|---|---|---|---|---|---|---|---|---|---|---|---|
| | T1 | U | S | H | T1 | U | S | H | T1 | U | S | H | T1 | U | S | H |
| ESZSL[47] | 53.9 | 12.6 | **63.8** | 21.0 | 58.2 | 2.4 | 70.1 | 4.6 | 58.6 | 5.9 | 77.8 | 11.0 | 38.3 | 2.4 | 70.1 | 4.6 |
| ALE[1] | 54.9 | 23.7 | 62.8 | 34.4 | 59.9 | 16.8 | 76.1 | 27.5 | 62.5 | 14.0 | 81.8 | 23.9 | 39.7 | 4.6 | 73.7 | 8.7 |
| SJE[2] | 53.9 | 23.5 | 59.2 | 33.6 | 65.6 | 11.3 | 74.6 | 19.6 | 61.9 | 8.0 | 73.9 | 14.4 | 35.2 | 3.7 | 55.7 | 6.9 |
| DEVISE[18] | 60.3 | 52.2 | 42.4 | 46.7 | - | - | - | - | 59.7 | 17.1 | 74.7 | 27.8 | 39.8 | 4.9 | 76.9 | 9.2 |
| Baseline18-2 [59] | 57.3 | 43.7 | 57.7 | 49.7 | 68.2 | 57.9 | 61.4 | 59.6 | 65.3 | 56.1 | 65.5 | 60.4 | 42.6 | **32.9** | 61.7 | 42.9 |
| CIZSL[13] | - | - | - | - | - | - | - | - | 60.1 | - | - | 19.1 | 43.8 | - | - | 24.0 |
| GDAN[23] | - | 39.3 | 66.7 | 49.5 | - | - | - | - | 59.1 | 32.1 | 67.5 | 43.5 | 40.4 | 30.4 | 75.0 | 43.4 |
| MLSE[11] | - | 22.3 | 71.6 | 34.0 | - | - | - | - | - | 23.8 | 83.2 | 37.0 | - | 12.7 | 74.3 | 21.7 |
| VSE-S[66] | - | 33.4 | 87.5 | 48.4 | - | - | - | - | - | 41.6 | 91.3 | 57.2 | - | 24.5 | 72.0 | 36.6 |
| CADA-VAE[51] | - | 51.6 | 53.5 | 52.4 | - | 57.3 | 72.8 | 64.1 | - | 55.8 | 75.0 | 63.9 | - | - | - | - |
| GMN[50]a | **64.3** | 56.1 | 54.3 | **55.2** | **71.9** | 61.1 | 71.3 | 65.8 | - | - | - | - | - | - | - | - |
| Baseline18* [67] | 57.0 | 56.3 | 40.4 | 47.1 | 58.8 | 57.4 | 88.6 | 69.7 | 62.8 | 57.6 | 89.1 | 69.9 | 41.1 | 21.2 | 78.5 | 33.4 |
| **Baseline18*+CKL** | 57.6 | 56.1 | 47.6 | 51.5 | 59.1 | 58.8 | 91.3 | 71.5 | **66.0** | 58.4 | 91.2 | 70.9 | 42.0 | 24.3 | **79.2** | 37.2 |
| **Baseline18*+CKL+TR** | 58.2 | **57.8** | 50.2 | 53.7 | 62.4 | **61.4** | **93.2** | **74.0** | 64.3 | **61.2** | **92.6** | **73.7** | **44.3** | 30.8 | 78.9 | **44.3** |

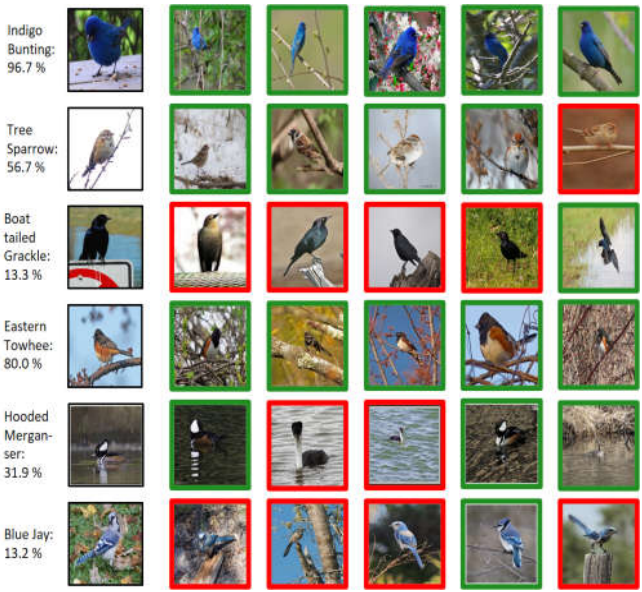

**Figure 3:** The representative ZSL retrieval results of the proposed method.

the proposed approach obtains up to 24.3%, 7.0%, 5.1% and 45.3% improvement in CUB, AwA1, AwA2 and aPY respectively. Compared with the state-of-the-art methods [13, 23, 11, 66, 51, 50], the proposed approach leads to significant improvements in term of H value on AwA1, AwA2 and aPY datasets. Specially, compared to CADA-VAE [51], our method achieves significantly higher H scores than CADA-VAE on AwA1 (74.0 to 64.1), AwA2 (73.7 to 63.9), and

CUB (53.7 to 52.4) datasets. And compared to GMN [50], our method significantly outperforms GMN on AwA1 dataset (74.0 to 65.8) but gets a little lower H scores on CUB dataset (53.7 to 55.2). In detail, our method has higher U scores (57.8 to 56.1) but relatively lower S scores (50.2 to 54.3) than GMN on CUB dataset. Because in our method, the terms of CKL and TR force the model to generate more generalized visual features for unseen classes recognizing that, to some extent, decrease the grain of visual features for seen classes. Since CUB dataset is all about fine-grained birds on visual features (like the detailed features about feathers, claws, wings, etc.), our method does not achieve the best H scores on the dataset. We have also tried to adjust the parameters to tradeoff the U and S score specifically on CUB dataset to get a higher H score (more than 55.0), but this cannot be extended to other datasets. In general, these results verified the outstanding generalization ability of our method and the proposed method outperforms the state-of-the-art methods on several datasets.

### 4.4. Hyperparameters Setting

There are three hyperparameters $\lambda_s, \lambda_g, \lambda_f$ in the loss function (Formula. 7). Fig. 4 shows the results of the proposed method with different hyperparameters on the CUB and NAB datasets. The x-axis represents family-level weights $\lambda_f$, the y-axis represents genus-level weights $\lambda_g$, and $\lambda_s$ is $1 - (\lambda_f + \lambda_g)$. The hyperparameters $\lambda_s, \lambda_g, \lambda_f$ are chosen from 0 to 1 (the step is 0.1) with the constraint $\lambda_s + \lambda_g + \lambda_f = 1$. The gray, orange and blue columns respectively represent that the Top-1 accuracy is less than Baseline18, less than the state-of-the-art (CIZSL), and greater the CIZSL. The proposed approach improves the Top-1 accuracy in most cases, especially in Fig. 4(d), which exceeds the state-of-the-art method (CIZSL) in most cases. It can be observed that the effect of hyperparameters on performance has the same trend depicted from the four sub-figures. When $\lambda_f$ or $\lambda_g$ is too large, the performance of our approach is reduced. This is because more generalized knowledge (from Family and Genus level) is introduced into the loss function, which reduces the weight to learn the specific knowledge (like from Species level). When $\lambda_f$ or $\lambda_g$ is too small, the performance of our approach is also been reduced. This is because more specialized knowledge (from Species level) is introduced into the loss function, which reduces the weight to learn the generalized knowledge (like from Family and Genus level). Thus, a suitable range of $\lambda_f$, $\lambda_g$ could be set respectively between 0.1-0.4 and 0.1-0.5 according to the result shown in Fig. 4.

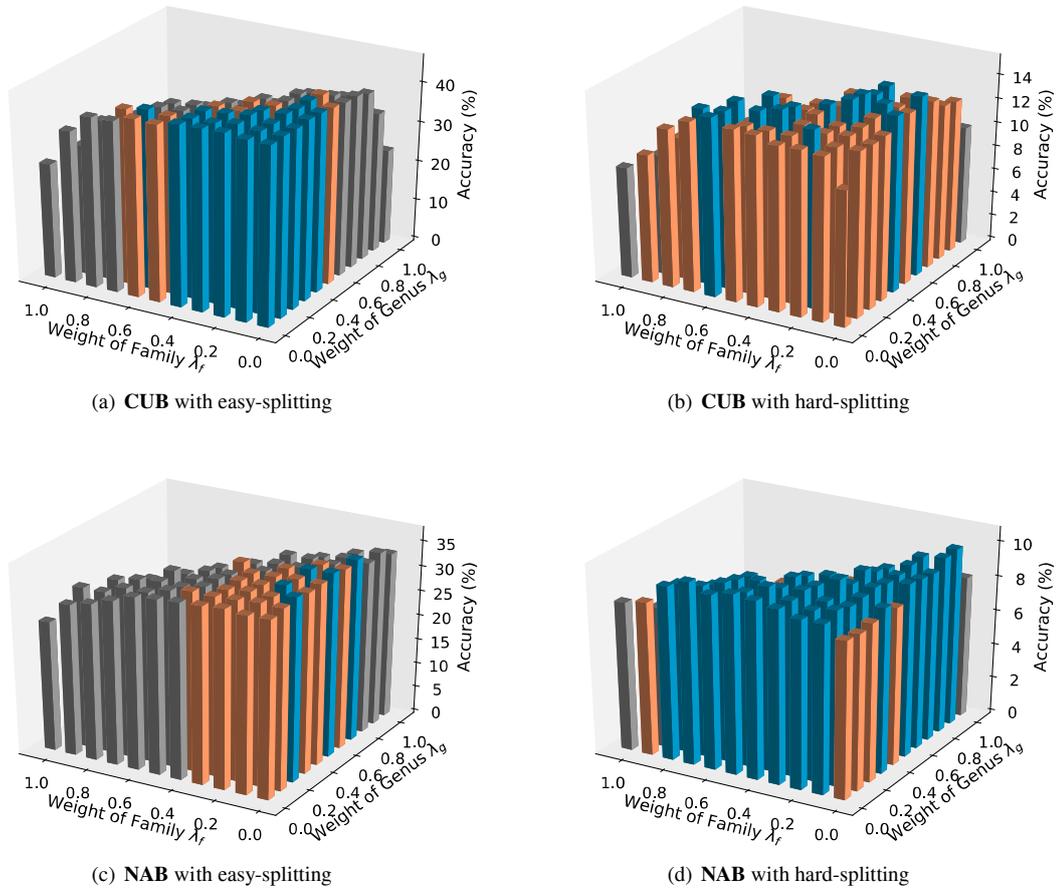

(a) **CUB** with easy-splitting

(b) **CUB** with hard-splitting

(c) **NAB** with easy-splitting

(d) **NAB** with hard-splitting

**Figure 4:** Top-1 accuracy of the proposed method with different hyperparameter settings. The gray, orange and blue column respectively represent that the Top-1 accuracy is less than Baseline18, less than the state-of-the-art (CIZSL), and greater the CIZSL.

### 4.5. Summary of The Experiment

From experimental results shown in Table 2, Table 3 and Table 4, the proposed method significantly outperforms the baseline methods and the state-of-the-art methods in all ZSL tasks, including ZSL image classification on CUB, NAB, AwA2 and aPY with different settings, ZSL image retrieval on CUB and NAB with different splittings. The experimental results indirectly confirm the proposed method can ef-fectively alleviate the problems of cross-modality and cross-domain. In details, we conducted two extra experiments to show why the proposed method meets the cross-modality and cross-domain challenges.

#### 4.5.1. Cross-Modality

In this work, we proposed so-called Cross-Knowledge Learning (CKL) to meet the cross-modality challenge. The idea of CKL is to introduce more semantic features to reduce the imbalance between semantic features and visual features. In the experiment, a taxonomy of birds and animals is introduced to bring more semantic features for each training sample. Fig. 5 provides an intuitive result to show how cross knowledge is introduced. In Fig. 5 (a), no cross knowledge is introduced, each image only uses its corresponding knowledge (Species). In Fig. 5 (b), cross knowledge is introduced by the Genus of the taxonomy, each image shares the knowledge with other images from the same Genus. In Fig. 5 (c), cross knowledge is introduced by the Family of the taxonomy, each image shares the knowledge with other images from the same Family.

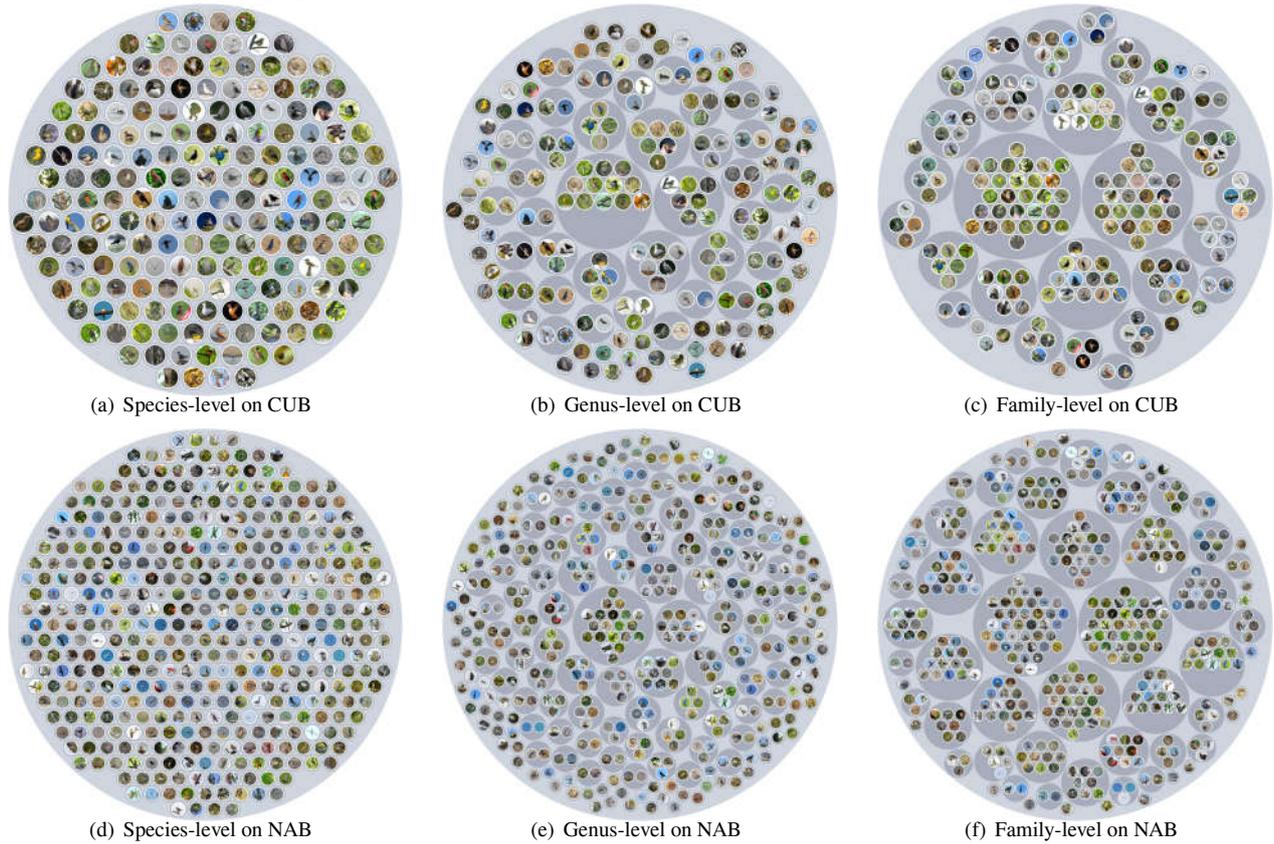

(a) Species-level on CUB     (b) Genus-level on CUB     (c) Family-level on CUB

(d) Species-level on NAB     (e) Genus-level on NAB     (f) Family-level on NAB

**Figure 5:** Cross knowledge is shared from Species, Genus, and Family of the taxonomy. (a) and (d) Each image only uses its corresponding knowledge (Species). (b) and (e) Each image shares the knowledge with other images from the same Genus. (c) and (f) Each image shares the knowledge with other images from the same Family

### 4.5.2. *Cross-Domain*

In this work, we proposed Taxonomy Regularization (TR) to meet the cross-domain challenge. The idea of TR is to regulate the loss function to guide the model to generate more generalized visual features according to the taxonomy. An extra experiment is conducted to visualize (t-SNE visualization) the synthesized visual features and the result is shown in Fig. 6. The circle represents the ground truth of visual features. The triangle represents synthesized visual features generated by the model. Different colors represent different classes. Fig. 6 (a) shows the synthesized visual features generated by Baseline17. The synthesized visual features of Baseline17 obviously deviate from the real visual features. Fig. 6 (b) shows the synthesized visual features generated by Baseline18. The synthesized visual features of Baseline18 do not intersect with the real visual features, but be surrounded by real visual features. It means the synthesized visual features are too specific compared with the real visual features. It is not good for the model to predict an unseen image if the synthesized visual features are too specific to the training set. Fig. 6 (c) shows the synthesized visual features generated by our method. It is obvious that the synthesized visual features of our method follow the distribution of the real visual features (Both the real and synthesized distribution of visual features have a hollow part.). Taxonomy Regularization forces the model to generate more general-ized visual features, which deviate from the corresponding seen classes, to adapt the unseen images in the prediction.

## 5. Conclusion

In this paper, we proposed a novel generative zero-shot learning approach to address the problems of cross-modality and cross-domain problems in ZSL, where Cross Knowledge Learning (CKL) is proposed to enrich the semantic features during the cross-modality learning, and Taxonomy Regularization (TR) is developed to encourage the generated samples to be able to deviate from the original class center to the center of more generalized classes for the cross-domain prediction. The experiment results demonstrate that our approach is superior to baselines and several state-of-the-art methods.

## Acknowledgment

The work is supported by the National Natural Science Foundation of China (Grant No. 61876166 and 61663046), and the Youth Talent Promotion Project of China Association for science and technology (Grant No. W8193209).

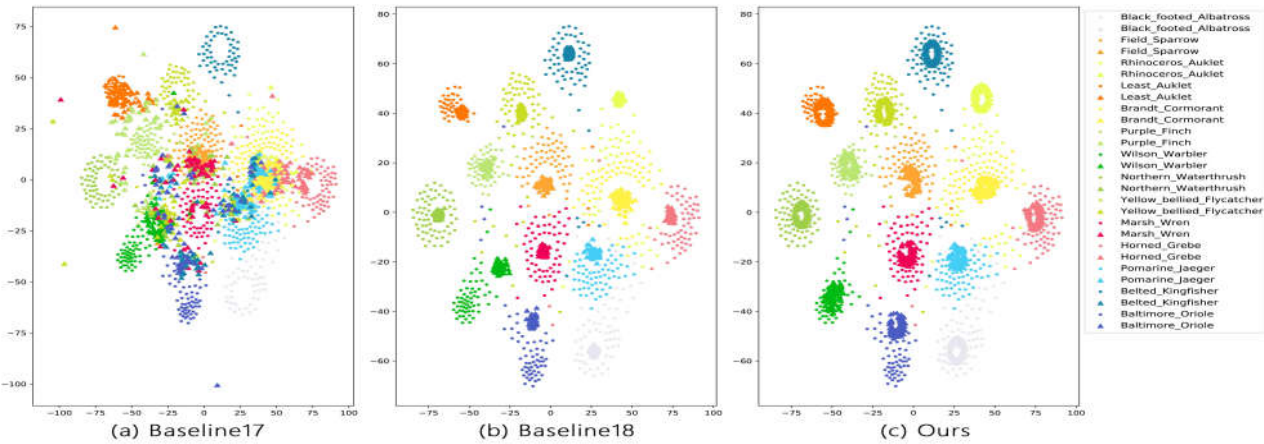

**Figure 6:** t-SNE visualization of real and synthesized visual features. The circle represents the real visual features. The triangle represents synthesized visual features. Different colors represent different classes.